\documentclass[conference]{IEEEtran}
\IEEEoverridecommandlockouts
\usepackage{cite}
\usepackage{amsmath,amssymb,amsfonts,amsthm}
\usepackage{algorithm}
\usepackage{algpseudocode}
\usepackage{graphicx}
\usepackage{textcomp}
\usepackage{xcolor}
\usepackage{url}
\usepackage{color}

\newcommand{\R}{\mathbb{R}}

\newcommand{\N}{\mathbb{N}}

\newcommand{\Prob}{\mathbb{P}}
\newcommand{\Exp}{\mathbb{E}}

\def\BibTeX{{\rm B\kern-.05em{\sc i\kern-.025em b}\kern-.08em
    T\kern-.1667em\lower.7ex\hbox{E}\kern-.125emX}}
\begin{document}

\title{Fast {Nonlinear} Risk Assessment for Autonomous Vehicles Using Learned {Conditional Probabilistic} Models of Agent Futures}

\newcommand{\revision}[1]{\textcolor{blue}{#1}}
\newcommand{\CH}[1]{\textcolor{red}{[CH: #1]}}
\newcommand{\AW}[1]{\textcolor{red}{[AW: #1]}}
\newcommand{\AJ}[1]{\textcolor{red}{[AJ: #1]}}
\let\oldemptyset\emptyset
\let\emptyset\varnothing

\author{Ashkan Jasour$^{*}$, Xin Huang$^{*}$, Allen Wang$^{*}$, and Brian C. Williams
\thanks{All authors are with the Computer Science and Artificial Intelligence Laboratory (CSAIL), Massachusetts Institute of Technology (MIT)
        {\tt\small \{jasour, huangxin, allenw, williams\} @ mit.edu}}
\thanks{*These authors contributed equally to the paper.}
       }

\maketitle

\begin{abstract}
This paper presents fast non-sampling based methods to assess the risk for trajectories of autonomous vehicles when probabilistic predictions of other agents' futures are generated by deep neural networks (DNNs). The presented methods address a wide range of representations for uncertain predictions including both Gaussian and non-Gaussian mixture models 
to predict both agent positions and control inputs {conditioned on the scene contexts}. We show that the problem of risk assessment when Gaussian mixture models (GMMs) of agent positions are learned can be solved rapidly to arbitrary levels of accuracy with existing numerical methods. To address the problem of risk assessment for non-Gaussian mixture models of agent position, we propose finding upper bounds on risk using {nonlinear} Chebyshev's Inequality and sums-of-squares (SOS) programming; they are both of interest as the former is much faster while the latter can be arbitrarily tight. These approaches only require higher order statistical moments of agent positions to determine upper bounds on risk. To perform risk assessment when models are learned for agent control inputs as opposed to positions, {we propagate the moments of uncertain control inputs through the nonlinear motion dynamics to obtain the exact moments of uncertain position over the planning horizon. To this end, we construct deterministic linear dynamical systems that govern the exact time evolution of the moments of uncertain position in the presence of uncertain control inputs.} The presented methods are demonstrated on realistic predictions from DNNs trained on the Argoverse and CARLA datasets and are shown to be effective for rapidly assessing the probability of low probability events.
\end{abstract}

\section{Introduction}
{Prediction under uncertainty plays a key role in safety of autonomous vehicles. More precisely,} in order for autonomous vehicles to drive safely on public roads, they need to predict the future states of other agents (e.g. human-driven vehicles, pedestrians, cyclists) and plan accordingly. Predictions, however, are inherently uncertain, so it is desirable to represent uncertainty in predictions of possible future states and reason about this uncertainty while planning. This desire is motivating ongoing work in the behavior prediction community to go beyond single mean average precision (MAP) prediction and develop methods for generating probabilistic predictions \cite{chai2019multipath, rhinehart2018r2p2, lee2017desire,huang2019diversity}. In the most general sense, this involves learning joint distributions for the future states of all the agents conditioned on their past trajectories and other context specific variables (e.g. an agent is at a stop light, lane geometry, the presence of pedestrians, etc). However, learning such a distribution can often be intractable, so current works use a wide variety of different simplified representations for probabilistic predictions. For example \cite{lee2017desire} trains a conditional Variational Autoencoder (CVAE) to generate samples of possible future trajectories. Other works use generative adversarial networks (GANs) to generate multiple trajectories with probabilities assigned to each of them \cite{huang2019diversity, li2019interaction}. As a discrete alternative, \cite{hong2019rules,bansal2018chauffeurnet} train a DNN to generate a probabilistic occupancy grid map with a probability assigned to each cell. However, such grid-based approaches effectively treat possible agents' trajectories as belonging to a discrete space, while, in reality, agents may be at an uncountable number of points in continuous space. Many recent papers try to account for the continuous nature of uncertainty in space by learning Gaussian mixture models (GMMs) for vehicle positions \cite{chai2019multipath, deo2018multi, hong2019rules} or coefficients of polynomials in $\R^2$ that represent the vehicles' positions \cite{huang2019uncertainty}. Since learning uncertain models for position or pose can also sometimes produce results that are inconsistent with basic kinematics, some recent works develop DNNs that predict future control inputs which are then propagated through a kinematic model to predict future positions \cite{cui2019deep, rhinehart2018r2p2}.

Given a probabilistic prediction, an autonomous vehicle still needs to be able to \textit{rapidly} evaluate the probability of a given plan resulting in a collision or, more generally, a constraint violation. We will refer to this problem as \textit{risk assessment} and it is particularly challenging in the context of autonomous driving as 1) autonomous vehicles need to reason about low probability events to be safer than human drivers and 2) there are hard real time constraints on algorithm latency. Latency is a critical consideration for safety and will be a major consideration motivating the methods presented in this paper. While an algorithm with a latency of, for example, one second would often be acceptable in other robotics applications, it would be unacceptable for an autonomous vehicle traveling at 20 m/s on public roads. This requirement of low latency while retaining the ability to reason about low probability events makes naive Monte Carlo computationally intractable. To address this problem, adaptive and importance sampling methods have been proposed to estimate these probabilities with fewer samples \cite{schmerling2016evaluating, norden2019efficient}. {However, such methods i) do not provide any bound on the risk and any safety guarantees, ii) are not suitable for real-time risk assessment without parallelization and running on GPU, and iii) lead to performance that is highly sensitive to algorithm parameters and proposal distributions.}

{Non-sampling based methods are widely used in risk assessment problems. Although existing non-sampling based methods are fast, they are limited to a particular class of uncertainties and safety constraints. For example, Gaussian-Linear methods use Boole's inequality }\cite{blackmore2011chance,blackmore2009convex,luders2010chance} {to estimate the probability of violation of linear constraints in the presence of Gaussian uncertainties. More precisely, the probability of a convex polytope is calculated as: 
$
\mbox{Prob} \{ \cap_{j=1}^{N} a_iX \leq b_j \} = 1- \mbox{Prob} \{ \cup_{j=1}^{N} a_iX
\geq b_j \}$ $\leq \sum_i \mbox{Prob} \{ a_iX \geq b_j \} )$. This results in a conservative upper bound on the risk. 
Chebyshev inequality based methods provide conservative bound on the risk of linear safety constraints using first two moments of uncertainties}\cite{summers2018distributionally,nakka2019trajectory}. {Conditional-value-at-risk (CVaR) based methods use a conservative approximation of the indicator functions of the safety constraints}, \cite{nemirovski2007convex,hong2011sequential}. { To evaluate the CVaR based risk constraints sampling-based methods or Gaussian uncertainties are used}, \cite{hakobyan2019risk,fan2021step}.

\indent\textit{Statement of Contributions}: We present fast methods to assess the risk of trajectories for both Gaussian and non-Gaussian position and control models of other agents. In section \ref{sec:risk_assess}, we begin by addressing the case when GMMs are used for agent position predictions. We show this particular case can be reduced to the problem of computing the CDF of a quadratic form in a multivariate Gaussian (QFMVG)- a well-studied problem in the statistics community for which methods exist that can rapidly solve it to arbitrary accuracy. To address the more general case when potentially non-Gaussian mixture models are used for agent position predictions, we apply statistical moment-based approaches to determine upper bounds on risk, which we will refer to as \textit{risk bounds}. 

\begin{figure}[!b]
    \vspace*{-5mm}
    \centering
    \includegraphics[width=0.8\linewidth]{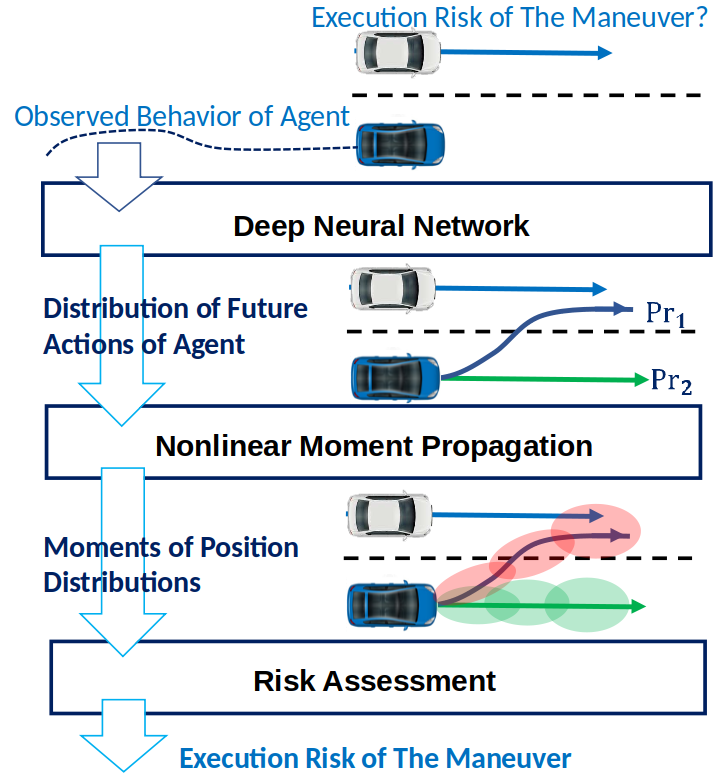}
    \caption{Illustration of our risk assessment framework when control distributions are used for predictions. When position distributions are used, the nonlinear moment propagation step is skipped.}
    \label{fig:control_framework_illustration}
\end{figure}

Namely, we propose using {nonlinear} Chebyshev's Inequality and a univariate sums-of-squares (SOS) program that can be seen as a generalization of Chebyshev's Inequality; the former is faster, while the latter can provide arbitrarily tight risk bounds. These moment-based approaches have the feature of being \textit{distributionally robust}, producing risk bounds that are true for all possible distributions that take on the value of the given moments.
To address uncertain models for control inputs, in Section \ref{sec:moment_prop}, {we propagate the moments of predicted probabilistic control inputs through the nonlinear motion dynamics to obtain the exact moments of uncertain position over the planing horizon. 
For this purpose, given the nonlinear motion dynamics and moments of uncertain control inputs, we construct new deterministic moment-state linear systems that govern the exact time evolution of the moments of uncertain position.} 
This enables the application of our non-Gaussian position risk assessment methods to the problem of risk assessment when models are learned for agent control inputs. Figure \ref{fig:control_framework_illustration} illustrates our framework in this case. {In Section~\ref{sec:predictor}, we present an encoder-decoder-based conditional predictor generating GMM predictions given the observed data and scene context.} In Section \ref{sec:experiments}, we demonstrate our methods on realistic predictions generated by DNNs trained on the Argoverse and CARLA datasets \cite{chang2019argoverse, Dosovitskiy17}. Source code can be found at \url{https://github.com/allen-adastra/risk_assess}.

\section{Notation}
Let $\mathcal{S}^n_{++}$ denote the set of $n\times n$ positive definite matrices. For any matrix $Q\in\mathcal{S}^n_{++}$ and vector $\mathbf{x}\in\R^n$, let $Q(\mathbf{x}):=\mathbf{x}^TQ\mathbf{x}$. Let $Q_{ij}$ denote the element in the $i_{th}$ row and $j_{th}$ column of $Q$. For any $\theta\in\R$, let $R(\theta)$ be the 2D rotation matrix parameterized by $\theta$. For a vector $\mathbf{x}\in\R^n$ and multi-index $\alpha\in\mathbb{N}^n$, let $x^\alpha = \prod_{i=1}^nx_i^{\alpha_i}$. For $n\in\N$, let $[n] = \{k\in\N : k\leq n\}$. 
For a vector valued function $f$, $f_i$ denotes the $i_{th}$ component of $f$.

\textbf{Polynomials:} Let {\small $\mathbb{R}[x]$} be the set of real polynomials in the variables {\small $x \in \mathbb{R}^n$}. Given polynomial
{\small$P(x):\mathbb{R}^n\rightarrow\mathbb{R}$}, we represent {\small $P$} as {\small $\sum_{\alpha\in\mathbb{N}^n} p_\alpha x^\alpha$} using the standard monomial basis {\small $\{x^\alpha\}_{\alpha\in \mathbb{N}^n}$} of $\mathbb{R}[x]$, and {\small $\mathbf{p}=\{p_\alpha\}_{\alpha\in\mathbb{N}^n}$} denotes the coefficients and $\alpha \in \mathbb N ^n$. Also, let $\mathbb R_{\rm d}[x] \subset \mathbb R [x]$ denotes the set of polynomials of degree at most $d\in \mathbb{N}$. 

\textbf{Sum of Squares Polynomials:}
 Polynomial {\small$P(x)$} is a sum of squares (SOS) polynomial if it can be written as a sum of \emph{finitely} many squared polynomials, i.e., {\small$P(x)= \sum_{j=1}^{m} h_j(x)^2$} for some $m<\infty$ and $h_j(x)\in\mathbb{R}[x]$ for $1\leq j\leq m$, 
SOS condition is a convex constraint that can be represented as a linear matrix inequality (LMI) in terms of coefficients of polynomial, i.e.,
{\small $P(x) \in SOS \rightarrow P(x)=\mathbf{x}TA\mathbf{x}$}, where $\mathbf{x}$ is the vector of standard basis and $A$ is a positive semidefinite matrix in terms of the coefficients of the polynomial, 

\textbf{Moments of Probability Distribution:}
For a random vector $\mathbf{w}$ and any $d\in\N$, let $\mu_{\mathbf{w}_t}, \Sigma_{\mathbf{w}_t}$ denote its mean vector and covariance matrix respectively, and $\Phi_\mathbf{w}$ denote its characteristic function.
Moments of random variables, are the generalization of mean and covariance and are defined as
expected values of monomials of random variables. More precisely, given $(\alpha_1,...,\alpha_n) \in \mathbb{N}^n$ where $\alpha=\sum_{i=1}^{n}\alpha_i$, moment of order $\alpha$ of random vector $\mathbf{w}$ is defined as $\mathbb{E}[ \Pi_{i=1}^n w_i^{\alpha_i}]$. Hence, the sequence of all moments of order $\alpha$ is defined as expected values of all monomials of order $\alpha$. For example, sequence of the moments of order $\alpha=2$ for $n=3$ is defined as 
\begin{small}
 $\left[\mathbb{E}[w_1^{2}], \mathbb{E}[w_1w_2],
\mathbb{E}[w_1w_3],
\mathbb{E}[w_2^2] ,
\mathbb{E}[w_2w_3],
\mathbb{E}[w_3^2] \right]$\end{small}.

Moment of order $\alpha$ can be computed by applying $n$ partial derivatives of the characteristic function as follows:

\begin{small}
 \begin{equation} \label{poly_mom_1}
    \mathbb{E}[ w_1^{\alpha_1}w_2^{\alpha_2}... w_n^{\alpha_n} ]= i^{-(\alpha_1+...+\alpha_n)} \frac{\partial^{\alpha_1+...+\alpha_n}}{\partial t_1^{\alpha_1}...\partial t_n^{\alpha_n}}  \Phi_{\mathbf{x}}(t_1,...,t_n)
\end{equation}
\end{small}
for $t_1=0,...,t_n=0$. {Note that for any probability density function, the characteristic function always exists} \cite{ccinlar2011probability}. {We will use  finite sequence of the moments to represent Non-Gaussian probability distributions.}

\section{Problem Statement}
We define \textit{risk} as the probability of an agent entering an ellipsoid around the ego vehicle. Thus, we are interested in computing the probability of other agents entering:
\begin{align}
    \left\{\mathbf{x}\in\R^2 : Q(\mathbf{x})\leq 1\right\}, \quad Q\in\mathcal{S}_{++}^2
\end{align}
We argue ellipsoids are a useful representation as: 1) they can be fit relatively tightly to the profiles of vehicles, and 2) the sizes of both the ego vehicle and agent can be accounted for by properly scaling the size of the ellipsoid around the vehicle. Throughout the paper, agent positions at each time step are always defined in the frame of the planned future poses of the ego vehicle unless stated otherwise; Section \ref{sec:change_frame} shows how moments of distributions can be expressed in different frames. Given this formulation, the ellipsoid is parameterized by a constant matrix $Q\in\mathcal{S}^2_{++}$ in the ego vehicle frame. In practice, multiple ellipsoids can be defined around the vehicle and an appropriate one selected at run-time. We restrict our focus to the single agent case and note that the risk in a multi-agent setting can be upper bounded by summing the risk associated with each agent.

If $\mathbf{x}_t = [x_t, y_t]^T$ is some random vector for the position of the agent at time $t$, then the risk associated with an agent across the whole $T$ step time horizon is:
\begin{align}
    \mathcal{R} &:= \Prob\left(\bigcup_{t=1}^{T}\left\{Q(\mathbf{x}_t)\leq 1\right\}\right)\label{eq:agent_risk}
\end{align}
 By the inclusion-exclusion principle, the probability $(\ref{eq:agent_risk})$ can be computed as the sum of the probabilities of the marginal events and the probabilities of all possible intersections of events:
\begin{align}
\mathcal{R} = \sum_{J\in\mathcal{P}([T])} (-1)^{|J|+1}\Prob\left(\bigcap_{j\in J} \{Q(\mathbf{x}_{j})\leq 1\}\right)
\end{align}
In many works, the random variables are assumed to be independent across time or can be made to be independent across time by conditioning on a discrete mode \cite{chai2019multipath,deo2018multi,hong2019rules}. If there is dependence across time, one would need the conditional distributions of the events which require additional information to be learned. As most work on behavior prediction currently assumes independence across time, this paper restricts its focus to the time independent case, and so:
\begin{align}
    \mathcal{R} = 1 - \prod_{t\in[T]} \left(1 - \Prob(Q(\mathbf{x}_t)\leq 1)\right)
\end{align}
Thus, the problem of risk assessment along the trajectory can be solved by computing the marginals at each time step $t$, so the rest of the paper restricts its focus to the marginals.
\section{Risk Assessment}\label{sec:risk_assess}
In this section, we present solutions for both Gaussian and non-Gaussian risk assessment when moments of the random vector for agent position $\mathbf{x}_t$ are known. We begin by addressing the problem of determining moments of agent positions in different frames to account for the ego vehicles planned trajectory. We then present our solution for the GMM case using numerical approximations of the CDFs of QFMVGs. To address the non-Gaussian case, we present methods based off Chebyshev's Inequality and SOS programming. We assume basic knowledge of SOS programming;
We refer the reader to \cite{parrilo2003semidefinite,rarnop} for an overview of SOS programming and \cite{jasour2019RiskMap,jasour2018moment,jasour2019Tube,jasour2016PhD,jasour2015SIAM,rarnop} for moment-SOS based planning under uncertainty.
Throughout this section, we assume the necessary moments of $\mathbf{x}_t$ are known.

\subsection{Changing Frames}\label{sec:change_frame}
Predictions are usually given in a global frame, so this section provides a method for transforming the global frame distribution moments into the ego vehicle frame. More generally, we are concerned with computing moments of $\mathbf{x}_t$ in a new frame offset by $\mathbf{v}\in\R^2$ and rotated by $-\theta\in\R$. As shown in the appendix, if $\mathbf{x}_t$ is a mixture model, its moments can be computed in terms of moments of its components, so, in this section, let $\mathbf{x}_t$ be a component of a mixture model. We propose only translating the moments and then accounting for the rotation by using $Q^* = R(\theta)^TQR(\theta)$ instead of $Q$. The rotation can be accounted for by using $Q^*$ instead of $Q$ because:
\begin{align}
    \mathbf{x}_t^TQ^*\mathbf{x}_t &= \mathbf{x}_t^TR(\theta)^TQR(\theta)\mathbf{x}_t\\&= (R(\theta)\mathbf{x}_t)^TQ(R(\theta)\mathbf{x}_t)
\end{align}
The translated moments can be computed by applying the binomial theorem to $(\mathbf{x}_t - \mathbf{v})^n$ (here, the power is applied element-wise). Note that applying the binomial theorem to $(\mathbf{x}_t - \mathbf{v})^n$ requires moments of $\mathbf{x}_t$ up to order $n$.

\subsection{Risk Assessment for GMM Position Models}
\begin{figure}[!t]
    \centering
    \includegraphics[width=0.95\linewidth]{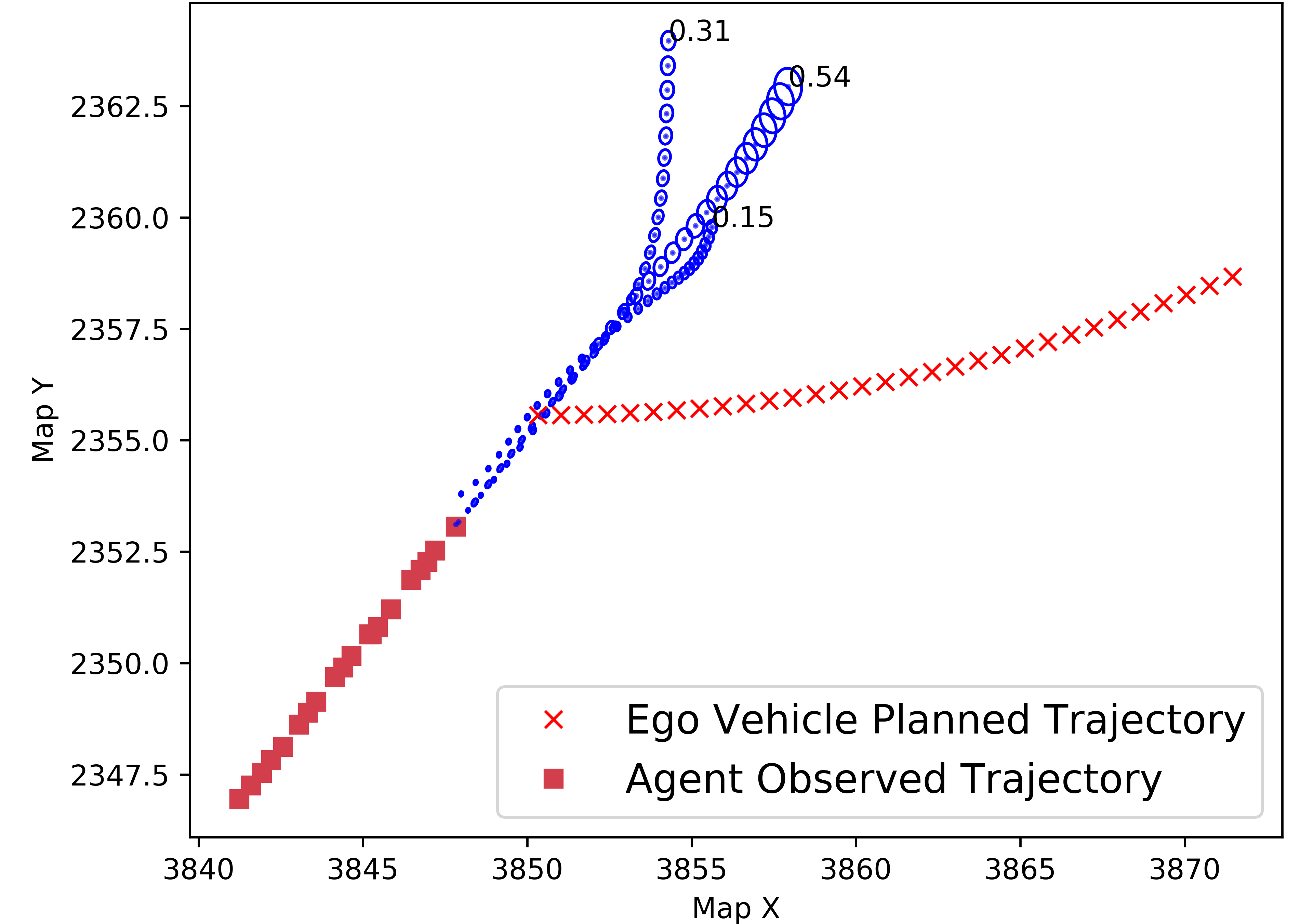}
    \caption{An example risk assessment scenario. One standard deviation confidence ellipses (in blue) of a multi-modal GMM prediction are shown with mode probabilities. The observed agent trajectory and planned ego vehicle trajectory are also shown in red with different markers.}
    \label{fig:risk_assess_example}
    \vspace*{-5mm}
\end{figure}

In this section, we provide a method to solve the risk assessment problem when the uncertain prediction is represented as a sequence of GMMs, $\mathbf{x}_t$, of the agents position with discrete modes determined by the Multinoulli $Z_t$. Many works currently learn GMMs for vehicle position as they express both multi-modal and continuous uncertainty \cite{chai2019multipath, deo2018multi, hong2019rules}. As shown in Figure~\ref{fig:risk_assess_example}, they provide an intuitive representation of uncertainty in both the drivers high level decisions and low level execution.
With time independence, the risk is:
\begin{align}
    \sum_{z = 1}^n\left(1 - \prod_{t\in[T]} 1 - \Prob(Q(\mathbf{x}_t)\leq 1 : Z_t=z)\right)\Prob(Z_t = z)
\end{align}
Note that the above expression can be easily modified for the case when there is a single Multinoulli random variable that is constant across all time, an assumption used in, for example, \cite{chai2019multipath}. The probabilities $\Prob(Z_t = z)$ are learned parameters of GMMs, so the problem of risk assessment can be solved by computing $\Prob\left(Q(\mathbf{x}_t)\leq 1 : Z_t = z\right)$ for each agent, time step, and mode. Note that this is exactly the CDF of $Q(\mathbf{x}_t)$ conditioned on $Z_t = z$ which is a quadratic form in a multivariate Gaussian (QFMVG). Unfortunately, there does not exist a known closed form solution to exactly evaluate the CDF of QFMVGs, but fast approximation methods with bounded errors have been studied within the statistics community \cite{liu2009new, solomon1977distribution, kotz1967series, duchesne2010computing, imhof1961computing}. Several of these methods have been implemented in the R package CompQuadForm \cite{de2017compquadform}. Of particular interest is the method of Imhof, which produces results with bounded approximation error by numerical inversion of the characteristic function of the QFMVG \cite{imhof1961computing}. A faster, but less accurate, alternative is the method of Liu-Tang-Zhang which involves approximating the CDF of the QFMVG with the CDF of a non-central chi square distribution with parameters chosen to minimize the difference in kurtosis and skew between the approximate and target distributions \cite{liu2009new}.
\subsection{Non-Gaussian Risk Assessment with Chebyshevs Inequality}
As a consequence of the one-tailed Chebyshev's Inequality, for any measurable function $g$, whenever $\mathbb{E}[g(\mathbf{x}_t)] > 0$, we have that:
\begin{align}
    \Prob(g(\mathbf{x}_t)\leq 0)&\leq \frac{\Exp[g(\mathbf{x}_t)^2] - \Exp[g(\mathbf{x}_t)]^2}{\Exp[g(\mathbf{x}_t)^2]}
\end{align}
That is, the first two moments of $g(\mathbf{x}_t)$ are sufficient to establish a bound on the risk that the constraint $g(\mathbf{x}_t)\leq 0$ is violated. We note that the requirement $\mathbb{E}[g(\mathbf{x}_t)] > 0$ is not particularly restrictive because $\mathbb{E}[g(\mathbf{x}_t)]\leq 0$ means the average case involves collision, thus corresponding to what is usually an unacceptable level of risk.
\subsubsection{Applying Chebyshev's Inequality to the Quadratic Form}\label{chebyshev_quad_form}
To apply Chebyshev's inequality to $\Prob(Q(\mathbf{x}_t) - 1\leq 0)$, we would need the first two moments of $Q(\mathbf{x}_t)-1$ which can be expressed in terms of the first two moments of $Q(\mathbf{x}_t)$. The first moment can be expressed in terms of the mean vector and covariance matrix of $\mathbf{x}_t$ \cite{provost1992quadratic}:
\begin{align}
    \mathbb{E}[Q(\mathbf{x}_t)] = \text{Tr}(Q\Sigma_{\mathbf{x}_t}) + \mu_{\mathbf{x}_t}^TQ\mu_{\mathbf{x}_t}
\end{align}
We can determine an expression for $\mathbb{E}[(Q(\mathbf{x}_t)^2]$ via an alternate representation for the quadratic form:
\begin{align}
    \mathbb{E}[Q(\mathbf{x}_t)^2] &= \sum_{(i,j,k,l)\in [2]^4}Q_{ij}Q_{kl}\mathbb{E}\left[x_{t_i}x_{t_j}x_{t_k}x_{t_l}\right]
\end{align}
Thus, to compute the second moment of $Q(\mathbf{x}_t)$, we would need the moments of $\mathbf{x}_t$ of order up to four.
\subsubsection{Conservative Approximation with Half-Spaces}
It's possible to reduce the order of the moments that need to be propagated to two by instead approximating the ellipsoid as the intersection of $n_h$ half-spaces parameterized by $\mathbf{a}_i\in\R^2$ and $b_i\in\R$. The approximated set is thus:
\begin{align}
\mathcal{X}_{Approx} = \cap_{i=1}^{n_h}\{\mathbf{x}\in\mathbb{R}^2 : \mathbf{a}_i^T\mathbf{x} + b_i\leq 0\}
\end{align}
Since the probability of any individual event is greater than the probability of the intersection of events, we have that:
\begin{align}
    \Prob(\cap_{i=1}^{n_h} \{\mathbf{a}_i^T\mathbf{x}_t + b_i\leq 0\})\leq \min_{i\in[n_h]} \Prob(\mathbf{a}_i^T\mathbf{x}_t + b_i \leq 0)
\end{align}
So if we determine an upper bound on the probability of each $\Prob(\mathbf{a}_i^T\mathbf{x}_t + b_i \leq 0)$ with Chebyshev's Inequality, the minimum of the Chebyshev bounds will be an upper bound on our risk. Since $\mathbf{a}_i^T\mathbf{x}_t + b_i$ is an affine transformation of $\mathbf{x}_t$, its mean and variance can be expressed with the mean vector and covariance matrix of $\mathbf{x}_t$.
\subsection{Non-Gaussian Risk Assessment with SOS Programming}\label{sos_subsection}
When tighter risk bounds are desired than those obtained via Chebyshev's Inequality, for any measurable function $g$,  an \textit{univariate} SOS program can be used to upper-bound $\Prob(g(\mathbf{x}_t)\leq 0)$ -- the SOS program is univariate in the sense that it searches for a polynomial in a single indeterminant, not in the sense that there is only one decision variable \cite{jasour2018moment}. The fact that the SOS program is univariate is significant because the key disadvantages of SOS, scalability and conservatism, are not as limiting for univariate SOS because: 1) the number of decision variables in the resulting SDP scales quadratically w.r.t. the order of the polynomial we are searching for and 2) the set of nonnegative univariate polynomials is equivalent to the set of univariate SOS polynomials, allowing univariate SOS to explore the full space of possible solutions.

We begin by noting that the probability of constraint violation is equivalent to the expectation of the indicator function of the sub-level set of $g$:
\begin{align}
    \Prob(g(\mathbf{x}_t)\leq 0) = \int_{ \{ \mathbf{x}_t: g(\mathbf{x}_t)\leq 0\}} pr(\mathbf{x}_t)d\mathbf{x}_t= \mathbb{E}[\mathbf{1}_{g(\mathbf{x}_t)\leq 0}]
\end{align}
where, $pr(\mathbf{x}_t)$ is probability density function of $\mathbf{x}_t$ and $\mathbf{1}_{g(\mathbf{x}_t)\leq 0}$ is the indicator function of the sub-level set of $g$ defined as $\mathbf{1}_{g(\mathbf{x}_t)\leq 0} = 1 $ if $x \in  \{ \mathbf{x}_t: g(\mathbf{x}_t)\leq 0\} $, and
0 otherwise.

The expectation of the indicator function, however, is not necessarily easily computable. To solve this problem, we find some polynomial with a more easily computable expectation that upper bounds the indicator function. If we can find some univariate polynomial, $p: \R\rightarrow\R$ of order $d$ in some indeterminant $x\in\R$ with coefficients $c_k, k = 0,...,d$ that upper bounds the indicator function, then clearly the following implication holds by substitution:
\begin{align} \label{obj_sos1}
    p(x):= \sum_{k=0}^d c_k x^k\geq \mathbf{1}_{x\leq 0}\Rightarrow \sum_{k=0}^d c_kg(\mathbf{x}_t)^k\geq\mathbf{1}_{g(\mathbf{x}_t)\leq 0}
\end{align}
Given the coefficients $c_k$, if we apply the expectation w.r.t. the density function of $\mathbf{x}_t$ to both sides, then we can reduce the problem of finding an upper bound on $\Prob(g(\mathbf{x}_t)\leq 0)$ to that of computing moments of the random variable $g(\mathbf{x}_t)$:
\begin{align}\label{obj_sos2}
    \sum_{k=0}^dc_k\mathbb{E}[g(\mathbf{x}_t)^k]\geq \mathbb{E}[\mathbf{1}_{g(\mathbf{x}_t)\leq 0}] = \Prob(g(\mathbf{x}_t)\leq 0)
\end{align}
where, $\mathbb{E}[g(\mathbf{x}_t)^k]$ is the moment of order $k$ of random variable $g(\mathbf{x}_t)$. The moments of $g(\mathbf{x}_t)$, in turn, are computable in terms of the moments of $\mathbf{x}_t$, i.e., $\mathbb{E}[\mathbf{x}_t^{\alpha}], \alpha \in \mathbb{N}$, by expanding out the polynomial power and applying the linearity of expectation, \cite{jasour2018moment}. For example, if $g(\mathbf{x}_t) = x_t^2 + y_t^2$, then:
\begin{align}
    \mathbb{E}[g(\mathbf{x}_t)^3] = \mathbb{E}[x_t^6] + 3\mathbb{E}[x_t^4y_t^2] + 3\mathbb{E}[x_t^2y_t^4] + \mathbb{E}[y_t^6]
\end{align}
The moments of $\mathbf{x}_t$ can be computed using the moment generating function as in \eqref{poly_mom_1}. In this section, we assume that we know the necessary moments of $\mathbf{x}_t$ to compute the moments $\mathbb{E}[g(\mathbf{x}_t)^k], \forall k\in[d]$. We also normalize the moments, as doing so improves the numerical conditioning of the problem \footnote{normalization is valid because $\Prob(X\leq 0) = \Prob(cX\leq 0)$ for $c>0$}.

Now consider the following univariate SOS program in the indeterminant $x$ which can search for upper bound polynomial indicator function, i.e., $p(x):= \sum_{k=0}^d c_k x^k\geq \mathbf{1}_{x\leq 0}$, which minimizes the upper bound on risk \cite{jasour2018moment}:
\begin{subequations}
\begin{alignat}{2}
    \min_{p, s_1, s_2}\quad &\sum_{k=0}^d c_k\mathbb{E}[g(\mathbf{x}_t)^k]\label{opt_obj}\\
    & p(x) - 1 = s_1(x) - xs_2(x) \label{non_neg_on_negative}\\
    & p(x), s_1(x), s_2(x) \quad SOS \label{con_sos}
\end{alignat}
\end{subequations}
If the order of the polynomial is chosen to be $d = 2n$ for some $n\in\N$, then we should have that $\text{deg}(s_1) = d$ and $\text{deg}(s_2) = d - 2$. If $d = 2n + 1$ for some $n\in\N$, then we should have that $\text{deg}(s_1) = 2n$ and $\text{deg}(s_2) = 2n$. Note that constraints (\ref{non_neg_on_negative}) and (\ref{con_sos})
are the nonnegativity constraints of the indicator function $\mathbf{1}_{x\leq 0}$, i.e., $\mathbf{1}_{x\leq 0}=1$ if $x\leq 0$, and otherwise $0$. 
More precisely, the constraint (\ref{non_neg_on_negative}) enforces:
\begin{align}
    p(x) \geq 1\quad \forall x\leq 0 \label{noneg_on_negative_condition}
\end{align}
Also, according to the constraint (\ref{con_sos}), $p(x)$ is constrained to be SOS; So it is globally nonnegative, i.e, $p(x) \geq 0$. 
Hence, polynomial $p(x)$ is an upper bound polynomial approximation of the indicator function, i.e., $p(x)\geq\mathbf{1}_{x\leq 0}, \forall x\in\R$. Thus, according to \eqref{obj_sos1} and \eqref{obj_sos2}, the optimal objective value of this SOS program yields an upper bound on $\Prob(g(\mathbf{x}_t)\leq 0)$.

Also, note that SOS optimization $(18)$ is a convex optimization. More precisely, it has a linear cost function in terms of the coefficients of the polynomial $p(x)$ and convex constraints in the form of linear matrix inequalities in terms of the coefficients of the polynomials $p(x),s_1(x)$, and $s_2(x)$. Hence, we can solve the SOS optimization efficiently using the off-the-shelf LMI optimization solvers.\\

\textit{Remark 1}: We can solve the optimization in $(18)$ to obtain the polynomial indicator function i.e., $p(x)= \sum_{k=0}^d c_k x^k$, in the offline step. Hence, we can compute the upper bound of the risk in terms of the obtained polynomial indicator function and moments of the uncertain states of the agent vehicle, in real-time, i.e., $\Prob(g(\mathbf{x}_t)\leq 0) \leq \mathbb{E}[p(g(\mathbf{x}_t))] =\sum_{k=0}^dc_k\mathbb{E}[g(\mathbf{x}_t)^k].$
For more information see \cite{jasour2018moment} and
\url{https://github.com/jasour/Nonlinear-Risk-Assessment} \\

\textit{Remark 2}: 
The provided non-Gaussian risk assessment approaches in Sections IV.C and IV.D are less conservative than the existing non-sampling based methods. This is because, we are using higher order moments of uncertainties and also tight polynomial approximation of the indicator functions of safety constraints. One can improve the risk assessment results by increasing the number of the moments in SOS programming based approach.

\section{Moment Propagation}\label{sec:moment_prop}
While directly learning distributions for agents future positions can be an effective strategy, one major disadvantage is it can produce physically unrealistic predictions. \cite{cui2019deep, rhinehart2018r2p2} address this by learning distributions for control inputs and then propagating samples through a kinematic model. While the Kalman filter and its variants, such as the extended and unscented Kalman filters, can be used to propagate mean and covariance, they are not exact and do not immediately apply to higher order moments \cite{wan2000unscented,kalman1961new,julier2004unscented}.

In this section, we provide an approach for nonlinear moment propagation that can, in principle, work for moments up to arbitrary order \cite{mom_prop, jasour2016PhD}. Given a nonlinear motion model and a random vector for control inputs, $\mathbf{w}_t$, this section is concerned with the problem of computing statistical moments of the uncertain position $\mathbf{x}_t$ s.t. the non-Gaussian risk assessment methods presented in Section \ref{sec:risk_assess} can be applied. More precisely, we are looking for the moments of uncertain position $(x_t,y_t)$ of the form $\mathbb{E}[x_t^\alpha y_t^\beta]$ where $\alpha,\beta\in\N$. We use a stochastic version of the discrete-time Dubin's car to both demonstrate the general approach and to address the problem of agent risk assessment:
\begin{subequations}
\begin{align}
    x_{t+1} &= x_t + v_t\cos(\theta_t)\\
    y_{t+1} &= y_t + v_t\sin(\theta_t)\\
    v_{t+1} &= v_t + w_{v_t}\\
    \theta_{t+1} &= \theta_t + w_{\theta_t}
\end{align}
\label{eq:dynamics}
\end{subequations}
Above, the control vector is $\mathbf{w}_t = [w_{v_t}, w_{\theta_t}]$ where $w_{v_t}$ and $w_{\theta_t}$ are random variables describing the agent's acceleration and steering at time $t$ and are assumed to be independent. $\mathbf{x}_t = [x_t, y_t]$ is the position of some reference point on the agent in a fixed frame, $v_t$ is its speed, and $\theta_t$ is the angle of its velocity vector with respect to the fixed frame. The time steps $\Delta t$ for discretization are omitted for brevity; the values of the variables can simply be scaled accordingly.

To obtain the moments of the uncertain position over the planning horizon, we will propagate the moments of uncertain control inputs $\mathbf{w}_t$  through  the  nonlinear stochastic Dubin's model. To this end, we construct deterministic linear dynamical systems, i.e., mapping between the moments at time $t+1$ and $t$, that govern the exact time evolution of the  moments  of  uncertain  position  in  the  presence  of  uncertain control inputs. By obtaining such dynamical systems in terms of the moments, we can recursively propagate the initial moments of the uncertain position over the planning horizon.

\subsection{Motivating Example}\label{sec_motiv}

To show how our moment propagation algorithm works, we begin by showing how the dynamics of the first order moment for the state $x_t$ in system $(\ref{eq:dynamics})$ can be found manually, i.e., mapping between $\mathbb{E}[x_{t+1}]$ and $\mathbb{E}[x_{t}]$. Our proposed algorithm is essentially an automated version of this process. By substituting the equations $(\ref{eq:dynamics})$ in and applying the linearity of expectation, we arrive at the dynamics of the  moment:
\begin{equation} \label{dyna_mom1}
\mathbb{E}[x_{t+1}]=\mathbb{E}[x_{t}]+\mathbb{E}[v_t cos(\theta_t)]    
\end{equation}
To complete the obtained mapping between $\mathbb{E}[x_{t+1}]$ and $\mathbb{E}[x_{t}]$, we need to compute the update rule for the term $\mathbb{E}[v_t cos(\theta_t)]$. By substituting the equations $(\ref{eq:dynamics})$ in and applying the linearity of expectation, we arrive at the update rule of the moment as:

\begin{small}
\begin{align}
\mathbb{E}[v_{t+1} & cos(\theta_{t+1})]  =   \notag \\
 & \mathbb{E}[cos(\omega_{\theta_t})]\mathbb{E}[v_tcos(\theta_t)]+\mathbb{E}[\omega_{v_t}]\mathbb{E}[cos(\omega_{\theta_t})]\mathbb{E}[cos(\theta_t)] \notag \\
& -\mathbb{E}[sin(\omega_{\theta_t})]\mathbb{E}[v_tsin(\theta_t)]-\mathbb{E}[\omega_{v_t}]\mathbb{E}[sin(\omega_{\theta_t})]\mathbb{E}[sin(\theta_t)]    \notag
\end{align}
\end{small}\noindent where, $\mathbb{E}[\omega_{v_t}]$ is the first order moment of uncertain control input $\omega_{v_t}$ and can be computed using it's characteristic function as in \eqref{poly_mom_1}. Also, $\mathbb{E}[cos(\omega_{\theta_t})]$ and $\mathbb{E}[sin(\omega_{\theta_t})]$ are the first order trigonometric moments of uncertain control input $\omega_{\theta_t}$ and can be computed using it's characteristic function as shown in Appendix.B. To complete the obtained update rule, we need to compute the update rule of the terms $\mathbb{E}[v_tsin(\theta_t)]$, $\mathbb{E}[cos(\theta_t)]$, and $\mathbb{E}[sin(\theta_t)]$. By doing so, we can complete the dynamics of the moment $\mathbb{E}[x_t]$ in \eqref{dyna_mom1} 
in terms of a set of slack moments $\mathbb{E}[v_tsin(\theta_t)]$, $\mathbb{E}[v_tcos(\theta_t)]$, $\mathbb{E}[cos(\theta_t)]$, and $\mathbb{E}[sin(\theta_t)]$. This will produces a closed form set of equations that can recursively compute $\mathbb{E}[x_{t+1}]$ in terms of the moments of the initial uncertain states and moments of uncertain control inputs at each time step. 

Similarly, we can obtain the dynamics of the higher order moments of the uncertain position, i.e., $\mathbb{E}[x^{\alpha}_{t}y^{\beta}_{t}]$. This process, however, is tedious and is easily subject to human error, especially for larger moment orders $\alpha, \beta$. To address these issues, in \cite{mom_prop2, mom_prop3}, we developed $\textit{TreeRing}$ algorithm that uses a dependency graph to identify all the slack moment states needed to construct the moment dynamical systems. 

In this paper, we use a different approach and provide a general framework to construct the moment dynamical systems. The main idea is to transform the nonlinear stochastic Dubbin's model in \eqref{eq:dynamics} into a equivalent new augmented
linear-state system. In this case, due to the linear relation of
the states of the augmented linear-state system at time $t$ and
$t+1$, moments of order $\alpha$ of the states at time $t+1$ can be described only in terms of the moments of order $\alpha$ of the states at time $t$. Hence, we do not need to look for a set of slack moment states as described in \eqref{dyna_mom1}.

\subsection{Equivalent Augmented Linear-State System}
In \cite{mom_prop}, we show that nonlinear stochastic motion dynamics can be transformed in to equivalent linear-state dynamical systems by introducing suitable new state variables. In this section,
we define such equivalent augmented linear-state system for stochastic nonlinear Dubin's model \eqref{eq:dynamics} as follows:
\begin{equation}\label{sys_aug}
\mathbf{x}_{aug_{t+1}}=A_t(\omega_{\theta_t}, \omega_{v_t})\mathbf{x}_{aug_t} \end{equation}
where $\mathbf{x}_{aug}=[ x,y,vcos(\theta), vsin(\theta), cos(\theta), sin(\theta)]^T$ is the augmented state vector in terms of the position $(x,y)$ and a set of nonlinear functions of the states of the original nonlinear model in \eqref{eq:dynamics}. Also, 
matrix $A_t(\omega_{\theta_t},\omega_{v_t})$ is defined only in terms of the nonlinear functions  of  the  uncertain  control inputs $\omega_{\theta_t}$ and $\omega_{v_t}$ as follows:
\begin{center}
    \resizebox{1\linewidth}{!}{%
$A_t(\omega_{\theta_t}, \omega_{v_t})=\begin{bmatrix}  1 & 0 &  1 & 0 & 0 & 0 \\ 1 & 0 &  0 & 1 & 0 & 0  \\
0 & 0 & cos( \omega_{\theta_t} ) & -sin(   \omega_{\theta_t}) &  \omega_{v_t}cos(  \omega_{\theta_t})  & - \omega_{v_t}sin(\omega_{\theta_t})\\ 
0 & 0 & sin( \omega_{\theta_t}) & cos(  \omega_{\theta_t}) & \omega_{v_t}sin(\omega_{\theta_t})  &  \omega_{v_t}cos(\omega_{\theta_t})\\
 0 & 0 & 0 & 0& cos(\omega_{\theta_t}) & -sin( \omega_{\theta_t})\\ 0 & 0 & 0 & 0 & sin( \omega_{\theta_t}) &
cos(\omega_{\theta_t})
 \\\end{bmatrix} $}
\end{center}
 Note that the obtained augmented system in \eqref{sys_aug} is linear in terms of the states $\mathbf{x}_{aug}$ and also equivalent to the original stochastic nonlinear Dubin's model in \eqref{eq:dynamics}. We   use   the obtained augmented   linear-state   system in \eqref{sys_aug}  to   obtain moment-state  linear dynamical systems that  govern  the exact time  evolution of the moments of the uncertain position states $(x,y)$ in nonlinear stochastic system \eqref{eq:dynamics}.

\subsection{Moment-State Linear Systems}

We define the moment-state linear systems for the obtained augmented linear-state system in \eqref{sys_aug} as follows \cite{mom_prop}:
\begin{equation} \label{sys_mom}
    \mathbb{E}[\mathbf{x}_{aug_{t+1}}^{\alpha}]=A_{{mom_{\alpha}}_t}\mathbb{E}[\mathbf{x}_{aug_{t}}^{\alpha}]
\end{equation}
where, $\mathbb{E}[\mathbf{x}_{aug_{t}}^{\alpha}]$ is the vector of all moments of order $\alpha$ of the augmented state vector $\mathbf{x}_{aug}$ and $A_{mom_{\alpha}}$ is a matrix in terms of the moments of the uncertain control inputs. To construct the moment-state linear system in \eqref{sys_mom}, we need the expected values of the monomials of order $\alpha$ of the vector $\mathbf{x}_{aug_{t+1}}$, e.g.,
$\mathbb{E}[x_{aug_{t+1}}^{\alpha}]$. Since $\mathbf{x}_{aug_{t+1}}$ is a linear function of $\mathbf{x}_{aug_t}$ as in \eqref{sys_aug}, we can describe the moments of order $\alpha$ of $\mathbf{x}_{aug_{t+1}}$ completely in terms of the moments of order $\alpha$ of $\mathbf{x}_{aug_t}$. Hence, we do not need to look for a set of slack moment states as described in the motivating example of Section \ref{sec_motiv}.

For example, we obtain the moment-state linear system of order $\alpha=1$ of the form $ \mathbb{E}[\mathbf{x}_{aug_{t+1}}]=A_{{mom_{1}}_t}\mathbb{E}[\mathbf{x}_{aug_{t}}]$ where
\begin{small}
\begin{align}
\mathbb{E}&[\mathbf{x}_{aug_t}]  = \notag \\
& \left[ \mathbb{E}[x_t], \mathbb{E}[y_t],  \mathbb{E}[v_tcos(\theta_t)], \mathbb{E}[v_tsin(\theta_t)],  \mathbb{E}[cos(\theta_t)],  \mathbb{E}[sin(\theta_t)]    \right ]^T \notag
\end{align}
\end{small}
is the vector of all moments of order $\alpha=1$ of $\mathbf{x}_{aug_t}$. 
Also, matrix $A_{{mom_{1}}_t}$ is described in terms the first order moments of uncertain control input $\omega_{v_t}$ and first order trigonometric moments of uncertain control input $\omega_{\theta_t}$ as follows:
\begin{center}
    \resizebox{1\linewidth}{!}{%
$A_{{mom_1}_t}=\begin{bmatrix}  1 & 0 &  1 & 0 & 0 & 0 \\ 1 & 0 &  0 & 1 & 0 & 0  \\
0 & 0 & \mathbb{E}[cos( \omega_{\theta_t} )] & -\mathbb{E}[sin(   \omega_{\theta_t})] &  \mathbb{E}[\omega_{v_t}]\mathbb{E}[cos(  \omega_{\theta_t})]  & - \mathbb{E}[\omega_{v_t}]\mathbb{E}[sin(\omega_{\theta_t})]\\ 
0 & 0 & \mathbb{E}[sin( \omega_{\theta_t})] & \mathbb{E}[cos(  \omega_{\theta_t})] & \mathbb{E}[\omega_{v_t}]\mathbb{E}[sin(\omega_{\theta_t})]  &  \mathbb{E}[\omega_{v_t}]\mathbb{E}[cos(\omega_{\theta_t})]\\
 0 & 0 & 0 & 0& \mathbb{E}[cos(\omega_{\theta_t})] & -\mathbb{E}[sin( \omega_{\theta_t})]\\ 0 & 0 & 0 & 0 & \mathbb{E}[sin( \omega_{\theta_t})] &
\mathbb{E}[cos(\omega_{\theta_t})]
 \\\end{bmatrix} $}
\end{center}
 We can compute the moments of uncertain control inputs in terms of the characteristic functions as shown in \eqref{poly_mom_1} and Appendix.B.
The obtained first order moment-state linear dynamical system describes the
exact time evolution of the first order moments of the uncertain position $(x,y)$ of the original stochastic nonlinear system in \eqref{eq:dynamics}. Similarly, we can obtain the moment-state linear system of the form \eqref{sys_mom} for higher moment order $\alpha$ to describe the exact time evolution of the higher order moments of the uncertain position.

Note that we can construct the deterministic linear moment-state systems of the form \eqref{sys_mom} for different moment order $\alpha$ in the offline step. Hence, we can use the obtained moment systems to propagate the moments of the uncertain initial states over the planning horizon in real-time. 
More precisely, given initial moments $\mathbb{E}[\mathbf{x}_{aug_0}^{\alpha}]$ and $A_{{mom_{\alpha}}_t}|_{t=0}^{N-1}$, the moments at time step $N$ can be obtained by recursion of  
$\mathbb{E}[\mathbf{x}_{aug_{t+1}}^{\alpha}]=A_{{mom_{\alpha}}_t}\mathbb{E}[\mathbf{x}_{aug_{t}}^{\alpha}], t=0,...,N-1$. Similarly, we can describe the moments by the solution of the linear moment system as  $\mathbb{E}[\mathbf{x}_{aug_{N}}^{\alpha}]=\Pi_{t=0}^{N-1} A_{{mom_{\alpha}}_t}\mathbb{E}[\mathbf{x}_{aug_{0}}^{\alpha}]$.\\

{
\textit{Remark 3}: The provided approach is not limited to the motion dynamics in $(20)$. We can use different motion dynamics with different sources of uncertainties to model the uncertain behaviour of the agent vehicles. For more information see} \cite{mom_prop} and \url{https://github.com/jasour/Uncertainty-Propagation}



\section{Deep Neural Network Predictor}\label{sec:predictor}

To obtain probabilistic future states of agents for risk assessment, we propose a conditional deep neural network that generates Gaussian mixture model parameters for \emph{positions} or \emph{controls} for a given sequence of observed positions over the past $20$ time steps and scene context. Although our framework works with any conditional prediction model that outputs GMM parameters for positions, we aim to generate accurate and realistic predictions by selecting an encoder-decoder-based predictor that utilizes long short-term memory (LSTM) units, because of the recent success of recurrent neural networks in trajectory prediction on different benchmarks \cite{alahi2016social,huang2019diversity,deo2018multi,cui2019deep}.

\subsection{Input}
\begin{figure}[t!]
    \centering
    \includegraphics[width=1.0\linewidth]{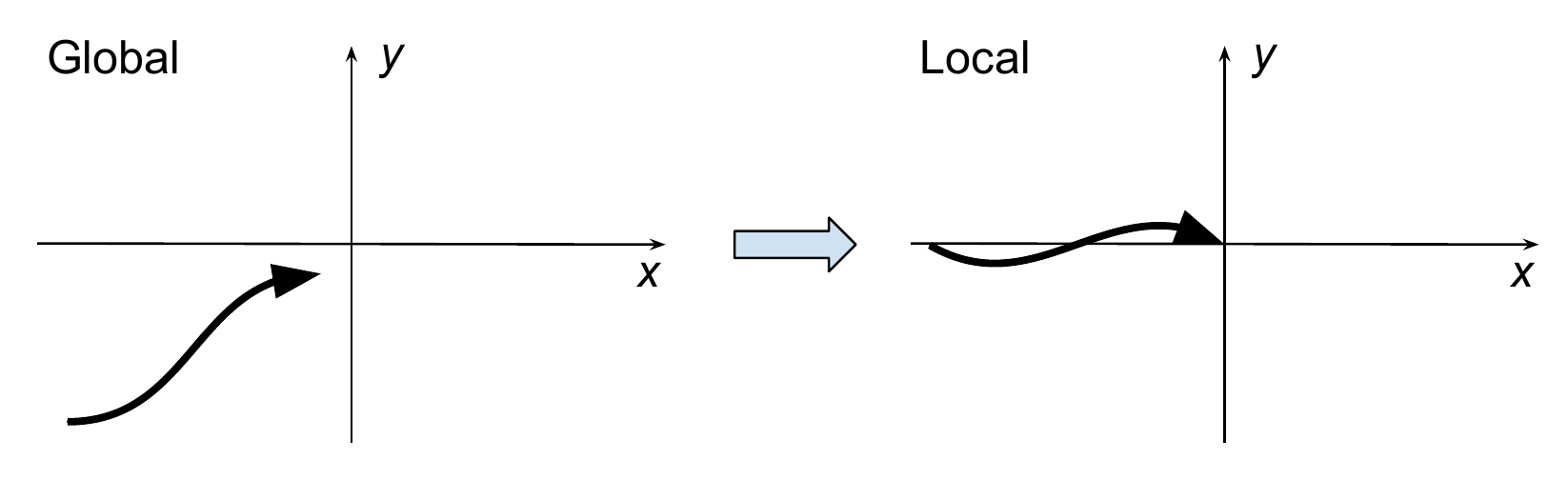}
    \caption{Illustration of trajectory normalization. Left: an observed trajectory in global coordinate. Right: an observed trajectory in its local coordinate, after normalization.}
    \label{fig:normalization}
\end{figure}

The input consists of observed vehicle positions $\mathbf{x}_{-19:0}$ for the past 20 timesteps, as well as scene context $\mathbf{c}$ representing driving context such as lanes coordinates $\mathbf{c}_M$ and the future trajectory of the ego car $\mathbf{c}_E$. Since the observed trajectory is usually collected in global coordinate, which can lead to bias in prediction, we normalize the past trajectory of the target vehicle, so that the last and first observed positions are at origin and the $x$-axis, respectively, as shown in Figure \ref{fig:normalization}.

The conditioning scene context is represented by a set of coordinates of the lanes that are close to the target vehicle (e.g., within 50 meters) and also the future planned trajectory of the ego car. The coordinates of the lanes and the ego car trajectory are normalized into the local coordinate of the target car. Such scene context provides additional cues on the possible future location of the target vehicle. Note that these contexts are usually not available across all driving platforms. Thus, we design a prediction model that is flexible in terms of the inputs, as we show in Section~\ref{sec:model}.

\subsection{Output}
The output $Y$ consists of a sequence of GMM parameters over the prediction horizon $T$ as follows:\\
$$Y = \{(w_t^{1}, \mu_t^{1}, \Sigma_t^{1}),\ldots,(w_t^{N}, \mu_t^{N}, \Sigma_t^{N})\}_{t=1}^T$$ 

\noindent where $N$ represents the number of components in the GMM and $w_t^i$ represents the weight of $i$th component at time step $t$ such that $\sum_{i=1}^N w_t^i = 1$. Also, mean $\mu$ and covariance $\Sigma$ represent the predicted uncertain position and predicted uncertain control inputs in 
GMM position predictor and GMM control predictor, respectively.

\subsection{Model Architecture}
\label{sec:model}
\begin{figure}[t!]
    \centering
    \vspace*{-5mm}
    \includegraphics[width=1.0\linewidth]{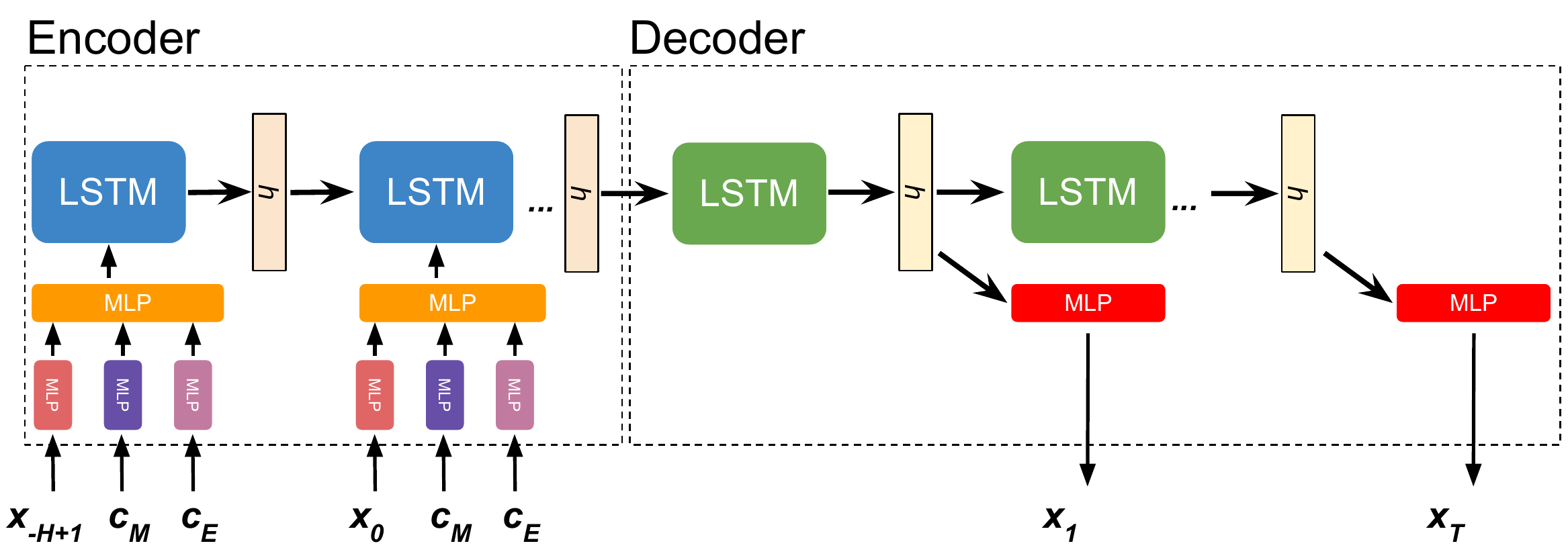}
    \caption{Architecture diagram of GMM predictor, including an LSTM-based encoder and an LSTM-based decoder. We introduce extra multilayer perceptrons (MLPs) to process the inputs in the encoder and process the predicted hidden states before generating predicted parameters in the decoder.}
    \label{fig:gmm_predictor}
\end{figure}

The encoder-decoder predictor is shown in Figure~\ref{fig:gmm_predictor}. The encoder is a sequence of LSTM units taking observed trajectories of the target agent and scene context as input and outputs a latent vector encoding agent hidden state. The input modalities are processed with separate multilayer perceptrions (MLPs) before being fed into the LSTM unit. This allows our model to handle different input options given their availability.
The decoder is also a sequence of LSTM units that takes the latent vector and generates a set of GMM parameters from each LSTM unit with MLPs. For simplicity, we use three component mixture models. For each component, we generate a weight value, a mean vector, and a covariance matrix representing uncertainties of predictions.

\subsection{Losses}
Given the prediction $Y$ and the groundtruth position or control values $\hat{Y}$, the loss function is composed of 2 terms as follows:
\begin{equation}
    \mathcal{L} = \alpha\mathcal{L}_{L2}(Y, \hat{Y}) + \mathcal{L}_{NLL}(Y, \hat{Y}),
\end{equation}
where $\mathcal{L}_{L2}$ measures the $L_2$ loss between the predicted mean values and ground truth observations over the future trajectories, $\mathcal{L}_{NLL}$ measures the negative log-likelihood loss between the predicted distributions and groundtruth observations, and $\alpha$ is the weight coefficient for the $L_2$ loss.

\section{Experiments}\label{sec:experiments}
In this section, we demonstrate the performance of our system through two learning-based predictors that predict stochastic position and control for the target agent. For each predictor, we describe its network details and training procedure, before presenting risk assessment results. All computations were performed on a desktop with an Intel Core i9-7980XE CPU at 2.60 GHz. All Monte Carlo (MC) methods are implemented with vectorized NumPy operations to have a realistic assessment of run times for naive MC.

\subsection{GMM Position Predictor}
\subsubsection{Training Details}
To obtain probabilistic trajectory distributions of agents, we trained an encoder-decoder DNN described in Section~\ref{sec:predictor}, with the following details. Each MLP in Encoder is a 2-layer MLP followed by a ReLU activation function, where each layer consists of 32 hidden units. The LSTM units in both Encoder and Decoder has a hidden size of 32 with one layer. The MLP in Decoder is a single layer MLP that produces the desired output parameters. The model is trained and validated on a subset of the Argoverse dataset \cite{chang2019argoverse}. During training, we use a batch size of 32, learning rate of 0.0001, and $\alpha=0.1$.

\subsubsection{Prediction Experiments}
In order to validate the prediction performance of our model, we perform an ablation study and compare our model with baseline models, as summarized in Table~\ref{tab:prediction}. The performance is evaluated over standard Argoverse trajectory forecasting metrics \cite{chang2019argoverse} such as minimum-of-N average displacement error (MoN ADE) and minimum-of-N final displacement error(MoN FDE), which measures the best prediction error over the entire prediction horizon and at the end of prediction horizon, respectively.

We first present prediction results from two physics-based baselines that assume constant velocity and constant acceleration when generating predictions. Although working well over short horizons such as a few seconds, physics-based models fail to generate accurate predictions over 3 seconds. Furthermore, we compare our model with other models whose inputs are i) unnormalized global trajectories, ii) normalized trajectories, iii) normalized trajectories and ego car trajectory, and iv) normalized trajectories and both map context and ego trajectory. We observe that normalizing target car's past trajectory improve prediction performance by 12.67\% in terms of MoN ADE metric. On the other hand, ego car's trajectory is not as helpful as map context in terms of improving the prediction error.

\begin{table}[]
\centering
\caption{Results from GMM position predictor and baselines over a prediction horizon of three seconds. For each predictor, three prediction samples are selected and the error of the best sample is reported.}
\begin{tabular}{|l|l|l|l|l|}
\hline
\textbf{Model} & \textbf{MoN ADE (m)} & \textbf{MoN FDE (m)} \\ \hline
Constant Velocity           & 1.75  & 3.96                                                                               \\ \hline
Constant Acceleration     & 3.17 & 8.12                                                                                \\ \hline
DNN Unnormalized           & 1.50  & 2.80                                                                                \\ \hline
DNN Normalized  & 1.31 & 2.64                                                          \\ \hline
DNN Normalized+Ego  & 1.30 & 2.62                                                          \\ \hline
DNN Normalized+Ego+Map & 1.22 & 2.49                                                         \\ \hline
\end{tabular}
\label{tab:prediction}
\end{table}


\subsubsection{Risk Assessment Experiments}
On a dataset of 500 scenarios similar to that shown in Figure~\ref{fig:risk_assess_example}, predictions were made and the risk was evaluated along a predefined trajectory for the ego vehicle as shown in Table \ref{tab:risk}. To evaluate QFMVG's, we tested both the methods of Imhof and Liu-Tang-Zhang. The methods proposed are much faster than naive Monte Carlo with far lower error.  The method of Imhof with an error tolerance of $10^{-10}$ was used as ground truth \cite{imhof1961computing}. Only $170$ scenarios were used for error computation as results from scenarios with computed ground truth errors within tolerances (i.e: $10^{-10}$) were neglected for error computation. We note that the method of Liu-Tang-Zhang empirically produces results with very small errors while being several times faster than the method of Imhof, which may prove useful in certain contexts.
\begin{table}[]
\caption{Risk evaluation results for 500 scenarios involving thirty time steps three-mode GMM predictions. Errors correspond to the time step with the maximum error.}
\begin{tabular}{|l|l|l|l|}
\hline
\textbf{Method} & \textbf{\begin{tabular}[c]{@{}l@{}}Mean Time \\ (ms)\end{tabular}} & \textbf{\begin{tabular}[c]{@{}l@{}}Mean Max. \\Absolute Error\end{tabular}} & \textbf{\begin{tabular}[c]{@{}l@{}}Mean Max. \\Relative Error\end{tabular}} \\ \hline
Imhof           & $91.21$                   & $0.0$                                                                               & $0.0$                                                                               \\ \hline
Liu-Tang-Zhang  & $26.67$                   & $2.7\times 10^{-6}$                                                                          & $2.3\times 10^{-4}$                                                                           \\ \hline
MC $10^4$ & 106.9                   & $6.7\times 10^{-4}$                                                                           & 0.38                                                                              \\ \hline
MC $5\times 10^4$ & 422.5                   & $2.7\times 10^{-4}$                                                                           & 0.13                                                                              \\ \hline
MC $10^5$ & 1329                    & $1.9\times 10^{-4}$                                                                           & 0.12                                                                              \\ \hline
\end{tabular}
\label{tab:risk}
\vspace*{-5mm}
\end{table}

\subsection{GMM Control Predictor}
\subsubsection{Training Details}
We use a similar DNN as the GMM position predictor, but the output becomes instead a set of GMM parameters for control signals defined in \eqref{eq:dynamics}. When training and validating our model, instead of using the Argoverse dataset which has noisy differentiated control data due to perception noise, we use our own data collected from a naturalistic driving simulator called CARLA \cite{Dosovitskiy17} that provides accurate ground truth control values. The model is trained and validated on 10k samples collected in CARLA.
\begin{figure}[!b]
    \centering
    \vspace*{-5mm}
    \includegraphics[width=0.9\linewidth]{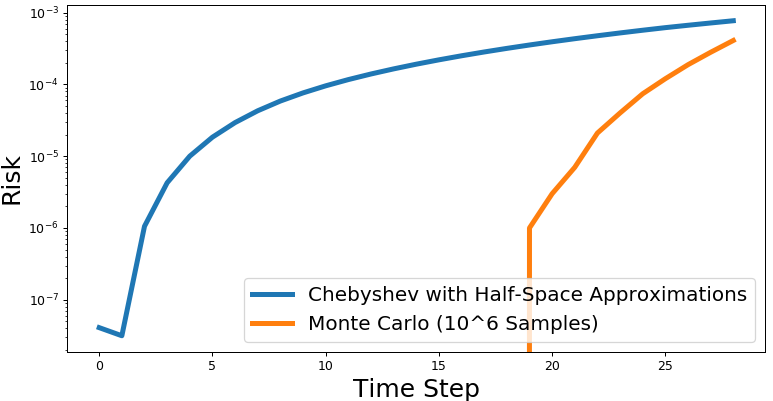}
    \caption{Risk estimates across time for an example scenario using random variables from the GMM control predictor.}
    \label{fig:uncertain_control_chebyshev}
\end{figure}
\subsubsection{Chebyshev Experiments} In this section the half-space approximation method with $12$ half-spaces was tested. The initial state of the agent vehicles was assumed to be known and deterministic. Random variables for control obtained using the DNN and moment-state dynamical systems were then used to compute the mean and covariance matrix of position at each time step. Over 50 scenarios, the mean time to evaluate the risk for a given trajectory for the Chebyshev method was 80ms while the Monte Carlo method with $10^6$ samples took 140 seconds. The average worst-case conservatism of the Chebyshev risk estimate for a given time step along a trajectory was $0.012$ (assuming the Monte Carlo results represent ground truth). Figure~\ref{fig:uncertain_control_chebyshev} shows the risk for both methods.

\subsubsection{Comparing SOS + Chebyshev}
\begin{figure}[!t]
    \centering
    \includegraphics[width=0.9\linewidth]{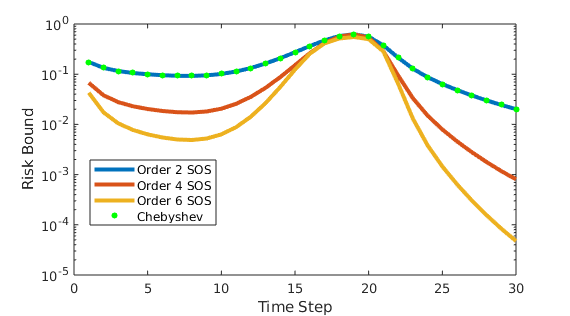}
    \caption{Risk bounds computed with the SOS formulation from section \ref{sos_subsection} compared with Chebyshev's inequality without half-space approximations.}\label{fig:my_label}
    \vspace*{-5mm}
\end{figure}

Experiments were run to test and compare the Chebyshev and SOS methods described in (\ref{chebyshev_quad_form}) and (\ref{sos_subsection}). For this experiment, higher order moments were obtained by automatic differentiation of the MVG moment generating function and the resulting moments of $Q(\mathbf{x}_t)-1$ were normalized. YALMIP was used to transcribe the SOS programs into Semidefinite programs, and SeDuMi was used to solve the resulting semidefinite programs \cite{Lofberg2004, sturm1999using}. As shown in Figure \ref{fig:my_label}, we observe that 1) Chebyshev bound produces nearly the same result as the second order SOS program and 2) the SOS program with higher order moments can yield significantly better bounds, especially in the tails. The solve times for each time step only marginally increased for the higher order SOS programs; the mean solve times were 42, 44, and 49 ms for the second, fourth, and sixth order SOS formulations, respectively. While these solve times obtained by solving univariate SOS programs are much better than those often encountered with multivariate SOS programs, further advances in performance are needed for this to be used online.



\section{Conclusions}
In this paper, we provided fast non-sampling based methods to propagate uncertainties and assess the risk for trajectories of autonomous vehicles when probabilistic predictions of other agents’ futures are generated by deep neural networks. Our experimental results with risk assessment methods for learned Gaussian mixture models of position and the Chebyshev and SOS program based risk assessment methods for learned non-Gaussian position and control models suggest that high performance implementations would be immediately practical for use in online applications.
As future work, we will incorporate uncertainty propagation and risk assessment into motion planning algorithms, but we note that it may be easily incorporated into standard algorithms with ``collision check" primitives such as RRTs and PRMs \cite{rarnop}. The SOS method does significantly improve upon the risk bounds from the Chebyshev method at the cost of additional computation; future research should further develop methods to make the SOS step offline to improve runtimes for online applications as proposed in \cite{jasour2018moment}. Future work in both prediction and risk assessment should also work towards relaxing assumptions such as time independence.

\section*{Acknowledgements}
This work was supported in part by Boeing grant MIT-BA-GTA-1 and by the Masdar Institute grant 6938857. Allen Wang was supported in part by a NSF Graduate Research Fellowship.

\section{Appendix}
\subsection{Moments and Characteristic Functions of Mixture Models}
Let $f_X$ denote the pdf of a $K$-component mixture model $X$, with pdf components $f_{X_i}, \forall i\in[K]$ and let $f_Z$ denote the pdf of the $K$ category Multinoulli. Then, by definition $f_X(x) = \sum_{i=1}^m f_{X_i}(x) f_Z(i)$.
For any measurable function $g$, by interchanging the order of integration and summation, the following holds true
\begin{align}
    \Exp[g(X)] &= \int g(x) f_X(x) dx\\
    &= \sum_{i=1}^K f_Z(i)\int g(x) f_{X_i}(x) dx
\end{align}
By letting $g(X) = X^n$ or $g(X) = e^{itX}$, the moments and characteristic function of $X$ can both be computed as the weighted sum of those of their components.
\subsection{Moments of Trigonometric Random Variables}
In this section, we show how trigonometric moments of the form $\mathbb{E}[\cos^n(X)]$, $\mathbb{E}[\sin^n(X)]$, and $\mathbb{E}[\cos^m(X)\sin^n(X)]$ can be computed in terms of the characteristic function of the random variable $X$, denoted by $\Phi_{X}$ \cite{mom_prop, mom_prop2}. We begin by applying Euler's Identity to the definition of the characteristic function as follows:
\begin{align}
\begin{split}
    \Phi_{X}(t) &= \mathbb{E}[e^{itX}]\\
    &= \mathbb{E}[\cos(tX)] + i\mathbb{E}[\sin(tX)]
\end{split}
\end{align}
Thus, we have that $\mathbb{E}[\cos(tX)] = \text{Re}(\Phi_{X}(t))$ and $\mathbb{E}[\sin(tX)] = \text{Im}(\Phi_{X}(t))$. This immediately gives us the ability to compute the first moments of our trigonometric random variables. For higher moments, the trigonometric power formulas can be used to express quantities of the form $\cos^n(X)$ as the sum of quantities of the form $\cos(mX)$ where $m\in\N$ and similarly for $\sin^n(X)$ \cite{zwillinger2002crc}. Thus, higher moments of $\sin(X)$ and $\cos(X)$ can be computed using 
$\Phi_X(t)$. More precisely, given $n \in \mathbb{N}, $ trigonometric moments of order $n$ of the forms $\mathbb{E}[{cos}^{n}(X)]$ and $  \mathbb{E}[{sin}^{n}(X)]$ reads as \cite{mom_prop}:
\begin{align} \label{Tri_mom_2}
\mathbb{E}[{cos}^{n}(X)] =\frac{1}{2^{n}} \sum_{k=0}^{n} \binom{n}{k}\Phi_{X}(2k-n)
\end{align}

\begin{align} \label{Tri_mom_3}
\mathbb{E}[{sin}^{n}(X)]&=\frac{(-i)^{n}}{2^{n}} \sum_{k=0}^{n} \binom{n}{k}(-1)^{n-k}\Phi_{X}(2k-n)
\end{align}
where, $\binom{n}{k}=\frac{n !}{k!(n-k)!}$. \\

Trigonometric moments of the form:
\begin{align}\label{eq:higher_cos_sin}
    \mathbb{E}[\cos^m(X)\sin^n(X)]
\end{align}
can also ultimately be computed in terms of $\Phi_X(t)$. This can be seen if we make the substitutions $\cos(X) = \frac{1}{2}(e^{ix} + e^{-ix})$ and $\sin(X) = \frac{1}{2i}(e^{ix} - e^{-ix})$, then (\ref{eq:higher_cos_sin}) can be expressed as:
\begin{align}
    \mathbb{E}\left[\frac{1}{i^n2^{m + n}}(e^{iX} + e^{-iX})^m(e^{iX} - e^{-iX})^n\right]
\end{align}
By applying the binomial theorem to  both expressions in parentheses, and multipying the resulting expressions, we find the entire expression in the expectation operator can be expressed as a polynomial in $e^{iX}$ and $e^{-iX}$. Thus, the entire expression can be written as the sum of terms of the form $\mathbb{E}[e^{itX}]$ for $t\in\mathbb{Z}$ which is in the definition of $\Phi_X(t)$. More precisely, given $(n,m) \in \mathbb{N}^2$, trigonometric moment of the form $\mathbb{E}\left[{cos}^{m}(X){sin}^{n}(X) \right]$ reads as \cite{mom_prop}:
\begin{equation} \label{Tri_mom_4}
\begin{split}
\hspace{-30mm} \mathbb{E}\left[{cos}^{m}(X){sin}^{n}(X) \right]=
\end{split}
\end{equation}

\resizebox{0.97\linewidth}{!}{%
$\frac{(-i)^{n}}{2^{m+n}} \sum_{(k_1,k_2)=(0,0)}^{(m,n)}  \binom{m}{k_1}\binom{n}{k_2}(-1)^{n-k_2}\Phi_{X}\left(2(k_1+k_2)-m-n\right) $}\\

\bibliographystyle{IEEEtran}
\bibliography{references} 

\begin{thebibliography}{10}
\providecommand{\url}[1]{#1}
\csname url@samestyle\endcsname
\providecommand{\newblock}{\relax}
\providecommand{\bibinfo}[2]{#2}
\providecommand{\BIBentrySTDinterwordspacing}{\spaceskip=0pt\relax}
\providecommand{\BIBentryALTinterwordstretchfactor}{4}
\providecommand{\BIBentryALTinterwordspacing}{\spaceskip=\fontdimen2\font plus
\BIBentryALTinterwordstretchfactor\fontdimen3\font minus
  \fontdimen4\font\relax}
\providecommand{\BIBforeignlanguage}[2]{{%
\expandafter\ifx\csname l@#1\endcsname\relax
\typeout{** WARNING: IEEEtran.bst: No hyphenation pattern has been}%
\typeout{** loaded for the language `#1'. Using the pattern for}%
\typeout{** the default language instead.}%
\else
\language=\csname l@#1\endcsname
\fi
#2}}
\providecommand{\BIBdecl}{\relax}
\BIBdecl

\bibitem{chai2019multipath}
Y.~Chai, B.~Sapp, M.~Bansal, and D.~Anguelov, ``Multipath: {M}ultiple
  probabilistic anchor trajectory hypotheses for behavior prediction,'' in
  \emph{Conference on Robot Learning (CoRL)}, 2019.

\bibitem{rhinehart2018r2p2}
N.~Rhinehart, K.~M. Kitani, and P.~Vernaza, ``{R2P2}: A reparameterized
  pushforward policy for diverse, precise generative path forecasting,'' in
  \emph{Proceedings of the European Conference on Computer Vision (ECCV)},
  2018, pp. 772--788.

\bibitem{lee2017desire}
N.~Lee, W.~Choi, P.~Vernaza, C.~B. Choy, P.~H. Torr, and M.~Chandraker,
  ``Desire: {D}istant future prediction in dynamic scenes with interacting
  agents,'' in \emph{Proceedings of the IEEE Conference on Computer Vision and
  Pattern Recognition}, 2017, pp. 336--345.

\bibitem{huang2019diversity}
X.~Huang, S.~G. McGill, J.~A. DeCastro, L.~Fletcher, J.~J. Leonard, B.~C.
  Williams, and G.~Rosman, ``Diversitygan: Diversity-aware vehicle motion
  prediction via latent semantic sampling,'' \emph{IEEE Robotics and Automation
  Letters}, vol.~5, no.~4, pp. 5089--5096, 2020.

\bibitem{li2019interaction}
J.~Li, H.~Ma, and M.~Tomizuka, ``Interaction-aware multi-agent tracking and
  probabilistic behavior prediction via adversarial learning,'' in \emph{2019
  International Conference on Robotics and Automation (ICRA)}.\hskip 1em plus
  0.5em minus 0.4em\relax IEEE, 2019, pp. 6658--6664.

\bibitem{hong2019rules}
J.~Hong, B.~Sapp, and J.~Philbin, ``Rules of the road: {P}redicting driving
  behavior with a convolutional model of semantic interactions,'' in
  \emph{Proceedings of the IEEE Conference on Computer Vision and Pattern
  Recognition}, 2019, pp. 8454--8462.

\bibitem{bansal2018chauffeurnet}
M.~Bansal, A.~Krizhevsky, and A.~Ogale, ``Chauffeurnet: Learning to drive by
  imitating the best and synthesizing the worst,'' in \emph{Robotics: Science
  and Systems}, 2019.

\bibitem{deo2018multi}
N.~Deo and M.~M. Trivedi, ``Multi-modal trajectory prediction of surrounding
  vehicles with maneuver based lstms,'' in \emph{2018 IEEE Intelligent Vehicles
  Symposium (IV)}.\hskip 1em plus 0.5em minus 0.4em\relax IEEE, 2018, pp.
  1179--1184.

\bibitem{huang2019uncertainty}
X.~Huang, S.~G. McGill, B.~C. Williams, L.~Fletcher, and G.~Rosman,
  ``Uncertainty-aware driver trajectory prediction at urban intersections,'' in
  \emph{2019 International Conference on Robotics and Automation (ICRA)}.\hskip
  1em plus 0.5em minus 0.4em\relax IEEE, 2019, pp. 9718--9724.

\bibitem{cui2019deep}
H.~Cui, T.~Nguyen, F.-C. Chou, T.-H. Lin, J.~Schneider, D.~Bradley, and
  N.~Djuric, ``Deep kinematic models for physically realistic prediction of
  vehicle trajectories,'' \emph{arXiv preprint arXiv:1908.00219}, 2019.

\bibitem{schmerling2016evaluating}
E.~Schmerling and M.~Pavone, ``Evaluating trajectory collision probability
  through adaptive importance sampling for safe motion planning,'' in
  \emph{Robotics: Science and Systems}, 2017.

\bibitem{norden2019efficient}
J.~Norden, M.~O'Kelly, and A.~Sinha, ``Efficient black-box assessment of
  autonomous vehicle safety,'' \emph{arXiv preprint arXiv:1912.03618}, 2019.

\bibitem{blackmore2011chance}
L.~Blackmore, M.~Ono, and B.~C. Williams, ``Chance-constrained optimal path
  planning with obstacles,'' \emph{IEEE Transactions on Robotics}, vol.~27,
  no.~6, pp. 1080--1094, 2011.

\bibitem{blackmore2009convex}
L.~Blackmore and M.~Ono, ``Convex chance constrained predictive control without
  sampling,'' in \emph{AIAA Guidance, Navigation, and Control Conference},
  2009, p. 5876.

\bibitem{luders2010chance}
B.~Luders, M.~Kothari, and J.~How, ``Chance constrained rrt for probabilistic
  robustness to environmental uncertainty,'' in \emph{AIAA guidance,
  navigation, and control conference}, 2010, p. 8160.

\bibitem{summers2018distributionally}
T.~Summers, ``Distributionally robust sampling-based motion planning under
  uncertainty,'' in \emph{2018 IEEE/RSJ International Conference on Intelligent
  Robots and Systems (IROS)}.\hskip 1em plus 0.5em minus 0.4em\relax IEEE,
  2018, pp. 6518--6523.

\bibitem{nakka2019trajectory}
Y.~K. Nakka and S.-J. Chung, ``Trajectory optimization for chance-constrained
  nonlinear stochastic systems,'' in \emph{2019 IEEE 58th Conference on
  Decision and Control (CDC)}.\hskip 1em plus 0.5em minus 0.4em\relax IEEE,
  2019, pp. 3811--3818.

\bibitem{nemirovski2007convex}
A.~Nemirovski and A.~Shapiro, ``Convex approximations of chance constrained
  programs,'' \emph{SIAM Journal on Optimization}, vol.~17, no.~4, pp.
  969--996, 2007.

\bibitem{hong2011sequential}
L.~J. Hong, Y.~Yang, and L.~Zhang, ``Sequential convex approximations to joint
  chance constrained programs: A monte carlo approach,'' \emph{Operations
  Research}, vol.~59, no.~3, pp. 617--630, 2011.

\bibitem{hakobyan2019risk}
A.~Hakobyan, G.~C. Kim, and I.~Yang, ``Risk-aware motion planning and control
  using cvar-constrained optimization,'' \emph{IEEE Robotics and Automation
  Letters}, vol.~4, no.~4, pp. 3924--3931, 2019.

\bibitem{fan2021step}
D.~D. Fan, K.~Otsu, Y.~Kubo, A.~Dixit, J.~Burdick, and A.-A. Agha-Mohammadi,
  ``Step: Stochastic traversability evaluation and planning for safe off-road
  navigation,'' \emph{arXiv preprint arXiv:2103.02828}, 2021.

\bibitem{chang2019argoverse}
M.-F. Chang, J.~Lambert, P.~Sangkloy, J.~Singh, S.~Bak, A.~Hartnett, D.~Wang,
  P.~Carr, S.~Lucey, D.~Ramanan \emph{et~al.}, ``Argoverse: 3d tracking and
  forecasting with rich maps,'' in \emph{Proceedings of the IEEE Conference on
  Computer Vision and Pattern Recognition}, 2019, pp. 8748--8757.

\bibitem{Dosovitskiy17}
A.~Dosovitskiy, G.~Ros, F.~Codevilla, A.~Lopez, and V.~Koltun, ``{CARLA}: {An}
  open urban driving simulator,'' in \emph{Proceedings of the 1st Annual
  Conference on Robot Learning}, 2017, pp. 1--16.

\bibitem{ccinlar2011probability}
E.~{\c{C}}{\i}nlar, \emph{Probability and stochastics}.\hskip 1em plus 0.5em
  minus 0.4em\relax Springer Science \& Business Media, 2011, vol. 261.

\bibitem{parrilo2003semidefinite}
P.~A. Parrilo, ``Semidefinite programming relaxations for semialgebraic
  problems,'' \emph{Mathematical programming}, vol.~96, no.~2, pp. 293--320,
  2003.

\bibitem{rarnop}
A.~Jasour, ``Risk aware and robust nonlinear planning,'' \emph{Course {N}otes
  for {MIT} 16.{S}498, rarnop.mit.edu}, 2019.

\bibitem{jasour2019RiskMap}
A.~Jasour and B.~C. Williams, ``Risk contours map for risk bounded motion
  planning under perception uncertainties,'' in \emph{2019 Robotics: Science
  and System (RSS)}, Germany, 2019.

\bibitem{jasour2018moment}
A.~Jasour, A.~Hofmann, and B.~C. Williams, ``Moment-sum-of-squares approach for
  fast risk estimation in uncertain environments,'' in \emph{2018 IEEE
  Conference on Decision and Control (CDC)}, 2445--2451, 2018.

\bibitem{jasour2019Tube}
A.~Jasour and B.~C. Williams, ``Sequential convex chance optimization for
  flow-tube based control of probabilistic nonlinear systems,'' in \emph{2019
  IEEE Conference on Decision and Control (CDC)}, France, 2019.

\bibitem{jasour2016PhD}
A.~Jasour, ``Convex approximation of chance constrained problems: Application
  in systems and control,'' in \emph{Dissertation in School of Electrical
  Engineering and Computer Science, The Pennsylvania State University}, 2016.

\bibitem{jasour2015SIAM}
A.~Jasour, N.~S. Aybat, and C.~M. Lagoa, ``Semidefinite programming for chance
  constrained optimization over semialgebraic sets,'' \emph{SIAM Journal on
  Optimization}, vol.~25, no.~3, pp. 1411--1440, 2015.

\bibitem{liu2009new}
H.~Liu, Y.~Tang, and H.~H. Zhang, ``A new chi-square approximation to the
  distribution of non-negative definite quadratic forms in non-central normal
  variables,'' \emph{Computational Statistics \& Data Analysis}, vol.~53,
  no.~4, pp. 853--856, 2009.

\bibitem{solomon1977distribution}
H.~Solomon and M.~A. Stephens, ``Distribution of a sum of weighted chi-square
  variables,'' \emph{Journal of the American Statistical Association}, vol.~72,
  no. 360a, pp. 881--885, 1977.

\bibitem{kotz1967series}
S.~Kotz, N.~L. Johnson, and D.~Boyd, ``Series representations of distributions
  of quadratic forms in normal variables {II}. {N}on-central case,'' \emph{The
  Annals of Mathematical Statistics}, vol.~38, no.~3, pp. 838--848, 1967.

\bibitem{duchesne2010computing}
P.~Duchesne and P.~L. De~Micheaux, ``Computing the distribution of quadratic
  forms: Further comparisons between the liu--tang--zhang approximation and
  exact methods,'' \emph{Computational Statistics \& Data Analysis}, vol.~54,
  no.~4, pp. 858--862, 2010.

\bibitem{imhof1961computing}
J.-P. Imhof, ``Computing the distribution of quadratic forms in normal
  variables,'' \emph{Biometrika}, vol.~48, no. 3/4, pp. 419--426, 1961.

\bibitem{de2017compquadform}
P.~L. De~Micheaux, ``Compquadform: distribution function of quadratic forms in
  normal variables,'' \emph{R package version}, vol.~1, no.~3, 2017.

\bibitem{provost1992quadratic}
S.~B. Provost and A.~Mathai, \emph{Quadratic forms in random variables: theory
  and applications}.\hskip 1em plus 0.5em minus 0.4em\relax M. Dekker, 1992.

\bibitem{wan2000unscented}
E.~A. Wan and R.~Van Der~Merwe, ``The unscented kalman filter for nonlinear
  estimation,'' in \emph{Proceedings of the IEEE 2000 Adaptive Systems for
  Signal Processing, Communications, and Control Symposium (Cat. No.
  00EX373)}.\hskip 1em plus 0.5em minus 0.4em\relax Ieee, 2000, pp. 153--158.

\bibitem{kalman1961new}
R.~E. Kalman and R.~S. Bucy, ``New results in linear filtering and prediction
  theory,'' 1961.

\bibitem{julier2004unscented}
S.~J. Julier and J.~K. Uhlmann, ``Unscented filtering and nonlinear
  estimation,'' \emph{Proceedings of the IEEE}, vol.~92, no.~3, pp. 401--422,
  2004.

\bibitem{mom_prop}
A.~Jasour, A.~Wang, and B.~C. Williams, ``Moment-based exact uncertainty
  propagation through nonlinear stochastic autonomous systems,''
  \emph{arXiv:2101.12490}, 2021.

\bibitem{mom_prop2}
A.~Wang, X.~Huang, A.~Jasour, , and B.~C. Williams, ``Fast risk assessment for
  autonomous vehicles using learned models of agent futures,'' \emph{Robotics:
  Science and System (RSS)}, 2020.

\bibitem{mom_prop3}
A.~Wang, A.~Jasour, , and B.~C. Williams, ``Non-gaussian chance-constrained
  trajectory planning for autonomous vehicles in the presence of uncertain
  agents,'' \emph{IEEE Robotics Automation Letters (RA-L)}, vol. 5(4), pp.
  6041–6048, 2020.

\bibitem{alahi2016social}
A.~Alahi, K.~Goel, V.~Ramanathan, A.~Robicquet, L.~Fei-Fei, and S.~Savarese,
  ``Social {LSTM}: Human trajectory prediction in crowded spaces,'' in
  \emph{Proceedings of the IEEE Conference on Computer Vision and Pattern
  Recognition}, 2016, pp. 961--971.

\bibitem{Lofberg2004}
J.~L{\"{o}}fberg, ``{YALMIP}: A toolbox for modeling and optimization in
  {MATLAB},'' in \emph{In Proceedings of the CACSD Conference}, Taipei, Taiwan,
  2004.

\bibitem{sturm1999using}
J.~F. Sturm, ``Using {SeDuMi} 1.02, a {MATLAB} toolbox for optimization over
  symmetric cones,'' \emph{Optimization methods and software}, vol.~11, no.
  1-4, pp. 625--653, 1999.

\bibitem{zwillinger2002crc}
D.~Zwillinger, \emph{{CRC} standard mathematical tables and formulae}.\hskip
  1em plus 0.5em minus 0.4em\relax Chapman and Hall/CRC, 2002.

\end{thebibliography}

\end{document}